\documentclass[10pt,twocolumn,letterpaper]{article}

\usepackage{iccv}              
\usepackage[accsupp]{axessibility}  

\usepackage{algorithm}
\usepackage{algorithmicx}  
\usepackage{algpseudocode}  
\usepackage{amsmath}  
\usepackage{multirow}
\usepackage{amssymb}
\usepackage{bbding}
\usepackage{array}
\usepackage{booktabs}
\usepackage{makecell}
\usepackage[table]{xcolor}

\definecolor{iccvblue}{rgb}{0.21,0.49,0.74}
\usepackage[pagebackref,breaklinks,colorlinks,allcolors=iccvblue]{hyperref}

\def\paperID{8523} 
\def\confName{ICCV}
\def\confYear{2025}

\title{SEGA: A Stepwise Evolution Paradigm for Content-Aware Layout Generation with Design Prior}

\author{Haoran Wang\\
Baidu Inc.\\
{\tt\small brucewang9111@gmail.com}
\and
Bo Zhao $^\dagger$ \\
Nanjing University\\
{\tt\small zhaobo20000116@gmail.com}
\and
Jinghui Wang\\
Baidu Inc.\\
\and
Hanzhang Wang\\
Harbin Institute of Technology\\
\and
Huan Yang\\
Kuaishou Technology\\
\and
Wei Ji \\
Nanjing University\\
\and
Hao Liu\\
Baidu Inc.\\
\and
Xinyan Xiao\\
Baidu Inc.\\
}

\begin{document}
\twocolumn[{
\maketitle
\vspace{-25pt}
\begin{center}
    \includegraphics[width=0.92\linewidth,height=6.3cm]{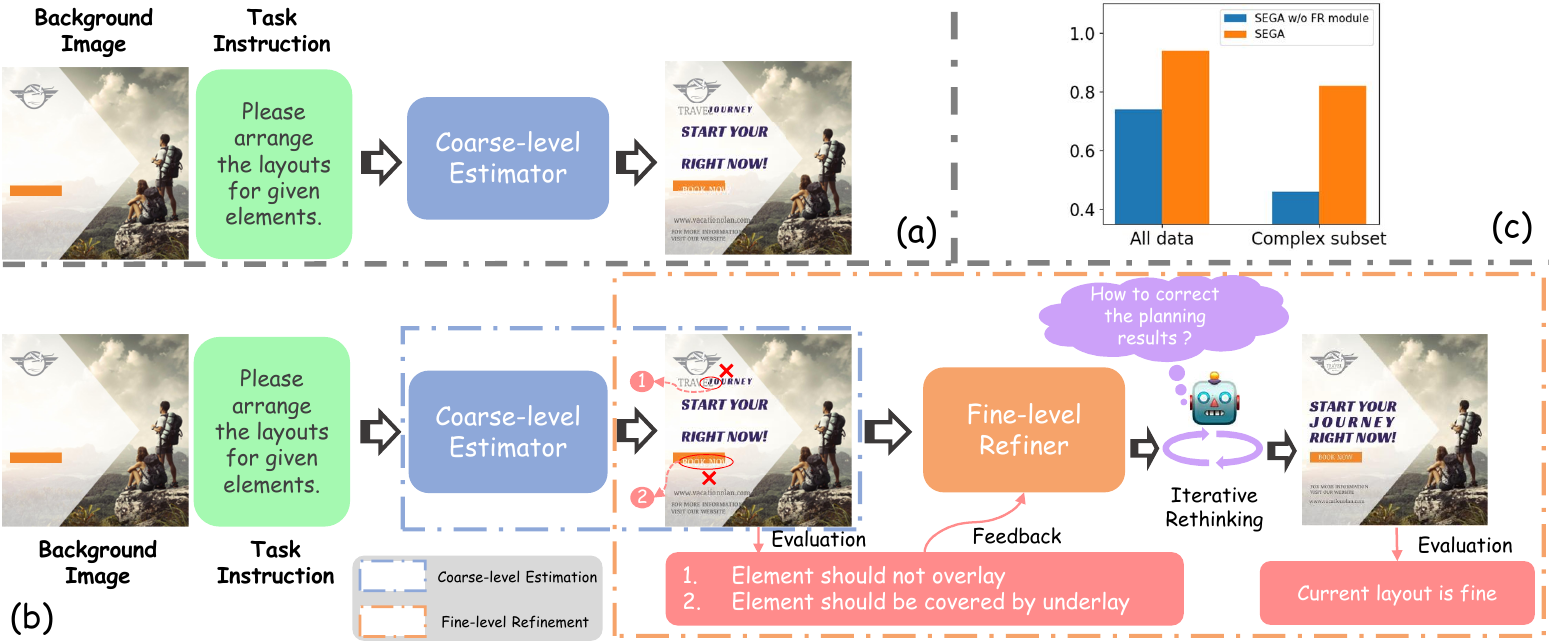}
    \captionof{figure}{Distinct from traditional methods using single-stage planning model (a), we propose a hierarchical layout generation paradigm (b) guided by design prior knowledge. Accordingly, our refinement module accepts an initial layout draft from the first stage and iteratively improves it. The superiority of our method is proven by handling difficult scenes on the Crello dataset (arranging $>$ 8 elements) (c). The metric in (c) is underlay accuracy which is introduced in the experiment section. }
    \label{fig:teaser}
\end{center}
}]
\renewcommand{\thefootnote}{}
\footnotetext{$\dagger$ Corresponding author.}

\begin{abstract}
In this paper, we study the content-aware layout generation problem, which aims to automatically generate layouts that are harmonious with a given background image. Existing methods usually deal with this task with a single-step reasoning framework. The lack of a feedback-based self-correction mechanism leads to their failure rates significantly increasing when faced with complex element layout planning. To address this challenge, we introduce \textbf{SEGA}, a novel \textbf{S}tepwise \textbf{E}volution Paradigm for Content-Aware Layout \textbf{G}ener\textbf{A}tion. Inspired by the systematic mode of human thinking, SEGA employs a hierarchical reasoning framework with a coarse-to-fine strategy: first, a coarse-level module roughly estimates the layout planning results; then, another refining module performs fine-level reasoning regarding the coarse planning results. Furthermore, we incorporate layout design principles as prior knowledge into the model to enhance its layout planning ability. Besides, we present GenPoster-100K that is a new large-scale poster dataset with rich meta-information annotation. The experiments demonstrate the effectiveness of our approach by achieving the state-of-the-art results on multiple benchmark datasets. Our project page is at: \url{https://brucew91.github.io/SEGA.github.io/}
\end{abstract}

\section{Introduction}
\label{sec:intro}

Layout generation aims at arranging design elements such as images, text, and logos to capture readers' attention and convey information in a visually harmonious manner with high aesthetic quality. Recently, it has attracted increasing attention in many real-scenario applications, such as designing posters~\cite{guo2021vinci}, web pages, and slides. Among various applications, content-aware layout generation~\cite{jyothi2019layoutvae,horita2024retrieval,seol2024posterllama,hsu2023posterlayout} is a hot-spot topic, which refers to generating a series of plausible location arrangements for elements, seamlessly blending with the provided background image.

To establish coherent graphic layouts, traditional studies typically train a specialized model, such as GAN~\cite{zhou2022cgl,hsu2023posterlayout} and Transformer~\cite{horita2023retrieval}, based on the encoder-decoder architecture~\cite{sutskever2014sequence}. Although attaining many achievements, these approaches are restricted by their reliance on predicting mere numerical values for layout design~\cite{tang2023layoutnuwa,seol2024posterllama}, consequently, failing to capture the semantic relationships among elements beyond the distribution of seen layouts. In contrast, the recent emergence of Large Language Models (LLM) ~\cite{achiam2023gpt,team2023gemini,touvron2023llama} has revolutionized numerous multi-modal vision-language comprehension tasks~\cite{wang2022coder,liu2023llava,bai2023qwen}. Correspondingly, PosterLlama~\cite{seol2024posterllama} exploits Llama~\cite{touvron2023llama} to reformat the layout elements arrangement task into code generation in an auto-aggressive manner, realizing apparent performance improvements. Overall, 
the commonality among currently prevailing methods lies in that they are all built based on the \textit{single-step learning paradigm}, in which the model receives input instructions and returns the final layout prediction through straightforward reasoning. 

Despite achieving steady advance, LLM's logical reasoning ability rooted in normative theories (like mathematics) still lags behind its formidable knowledge integration and memory abilities~\cite{mirzadeh2024gsm}. Specifically for graphic layout generation, we observe that its planning ability decreases significantly as the number of given elements becomes large. In contrast to neural networks, humans possess a structured cognitive system, which enables us to break down intricate problems into manageable sub-problems via
logical step-by-step thinking. Moreover, another critical learning mechanism of us humans is to rethink and learn from previous trial and error to avoid making similar mistakes again~\cite{mercer2008talk}. Intuitively, \textit{introducing a human-like thinking pattern is reasonable to enhance the model's reasoning ability}.

Inspired by this concept, we draw inspiration from the human cognitive system and propose a novel learning paradigm for content-aware layout generation. For humans, two families of cognitive operations, called fast thinking and slow thinking systems, are combined in decision-making~\cite{tay2016systems}. The former is adept at quickly solving intuitive and simple tasks, while the latter is responsible for tackling analytical and complex problems~\cite{kahneman2011thinking}. To mimic this thinking mechanism, we design a stepwise evolution learning paradigm (see Figure~\ref{fig:teaser}~(b)), which decomposes the layout generation into two stages: \textit{\textbf{coarse-level layout estimation}} and \textit{\textbf{fine-level layout refinement}}. In the first stage, we follow traditional methods to feed the input instructions into the \textit{\textbf{Coarse-level Estimation module}} (\textit{CE module}) and roughly estimate the intermediate layouts. In the second stage, we aim to obtain refined results with higher quality based on the intermediate layouts. Concretely, the \textit{\textbf{Fine-level Refinement module}} (\textit{FR module}) module is initialized by the CE module weights, indicating the former evolves from the latter. Then, we collect the intermediate results of the CE module on the training data and employ them plus additional perturbed layouts for FR module training, supervised by the pruned Ground Truth annotations. Analogously, the CE module acts like the quick-thinking system of humans, furnishing basic cues for assisting in-depth reasoning of the FR module. Experimental results prove our paradigm substantially boosts the model's planning ability for the complex layout. (see Figure~\ref{fig:teaser}~(c))

Another merit of human thought is our capacity to conduct logical inference through an in-depth understanding of internal concepts. Correspondingly, some works inspired by this mechanism have been proven to enhance the reasoning capabilities of LLMs. such as CoT~\cite{wei2022chain}. In this paper, we contribute to empowering LLMs with this strategy to improve graphic layout design ability. To avoid being misled by superficial numerical relationships, we select explicit design principles as cues to ignite the module's reasoning potential. Specifically, we extract the common sense of layout design into four principles, as illustrated in Figure~\ref{fig:teaser}~(b). Next, as output, our model first evaluates the current layout results and find out its design flaws according to the principles before the layout results are predicted. Driven by the mechanism of auto-regressive generation, the evaluation results are injected into the training procedure of the RF module. It enforces our model not only to predict the layout results but also to identify its drawbacks.

Furthermore, existing datasets for graphic layout tasks mostly rely on inpainting techniques to obtain background images, thus degrading their data quality. To advance the development of relevant fields, we contribute a new dataset named GenPoster-100K. It features high-fidelity component derived from layer-parseable source materials, encompassing over 100,000 diverse posters annotated with multi-grained metadata. 
Besides, We evaluate the cross-dataset generalization capabilities of models trained on our GenPoster-100K dataset when applied to other benchmarks.

Briefly, our contributions are summarized as follows: (\textit{\romannumeral 1}) We propose a novel \textbf{S}tepwise \textbf{E}volution Paradigm for Content-Aware Layout \textbf{G}ener\textbf{A}tion (\textbf{SEGA}), which leverages coarse-to-fine learning strategy and explicit knowledge feedback to boost the planning ability of Multi-modal Large language Models. (\textit{\romannumeral 2}) We curate a new large-scale and high-quality layout generation dataset named \textbf{GenPoster-100K}, which contains more than 100,000 diverse graphic layout samples and provides complete structured information for each component. (\textit{\romannumeral 3}) Extensive experiments not only verify that our method achieves substantial performance gains over state-of-the-art (SoTA) approaches on three benchmark datasets, but also exhibits satisfactory generalization ability brought by GenPoster-100K.

\section{Related work}

\subsection{Graphic Layout Generation}
The existing approaches about layout generation can be categorized into two types: 1) \textbf{content-agnostic} layout generation methods and 2) \textbf{content-aware} layout generation methods. The content-agnostic layout generation aims at designing layouts with no reference to given visual content, like a canvas. In the early stage, some methods attempted to employ a set of rules~\cite{o2014learning,cao2012automatic} or pre-defined templates~\cite{qian2020retrieve} to generate layouts. With the renaissance of deep learning, a line of methods introduced different deep architectures including VAE~\cite{jyothi2019layoutvae}, GAN~\cite{li2019layoutgan}, transformers~\cite{kong2022blt} and Diffusion models~\cite{inoue2023layoutdm,zhang2023layoutdiffusion} to solve the layout generation problem. Content-aware layout generation is designed to arrange spatial space for pre-defined elements on a given canvas. As a seminal work, ContentGAN~\cite{zheng2019content} introduced both visual and textual semantics into the generation of magazine layouts. More subsequent studies~\cite{cao2022geometry,zhou2022cgl,hsu2023posterlayout} chose to employ encoder-decoder architecture for visual contents-aware layout generation. Afterward, RALF~\cite{horita2023retrieval} applies retrieval augmentation to improve the layout generation ability via feeding the nearest neighbor layout examples into an auto-regressive generator. 

Large Language Models (LLMs)~\cite{achiam2023gpt,team2023gemini,touvron2023llama} have recently emerged and brought about a transformative impact on numerous applications~\cite{wu2023bloomberggpt,thirunavukarasu2023large,menon2022visual,yang2024vip}. Recent work shows LLMs' potential in layout generation. Tang et al.~\cite{tang2023layoutnuwa} transformed layout generation into code generation and tuned LLMs for it. Lin et al. \cite{lin2024layoutprompter} proposed a training-free in-context learning method to activate LLMs' layout design ability. Chen et al.~\cite{chen2023reason} used LLMs with Chain - of - Thought (CoT) prompting to convert captions to layout information. Seol et al.~\cite{seol2024posterllama} developed a two-stage training method for graphic layout generation, integrating a visual encoder and training a visual adapter in the second stage. 
Although making thrilling progress, the existing methods leverage monolithic Multi-modal Large Language Models (MLLMs) to perform single-stage layout generation. Distinct from the above approaches, we resolve this task by introducing a hierarchical learning paradigm, which is constructed by a coarse-level layout estimator along with a fine-grained layout refiner evolved from the former.

\subsection{Datasets for Graphic Layout Generation}
For poster layout generation, Zhou et al. contributed a layout dataset named CGL-Dataset~\cite{zhou2022cgl}, which contains 60,548 advertising posters with annotated layout information. The graphic elements are classified into predefined categories. Li et al. extended CGL-Dataset to a new CGL-Dataset V2~\cite{li2023relation}, where the content annotations of graphic elements are additionally involved. In comparison to CGL-Dataset and CGL-Dataset V2, Hsu et al. presented a median-scale dataset~\cite{hsu2023posterlayout} called PKU PosterLayout that covers a wider range of poster categories and supplies more complex layouts. Different from these datasets, Yamaguchi proposed the Crello dataset~\cite{yamaguchi2021canvasvae} which contains 23,182 posters collected from the web. Crello preserves all the separated elements for each poster.

 \begin{figure*}[t]
    \begin{center}
        \includegraphics[width=0.95\linewidth,height=6.2cm]{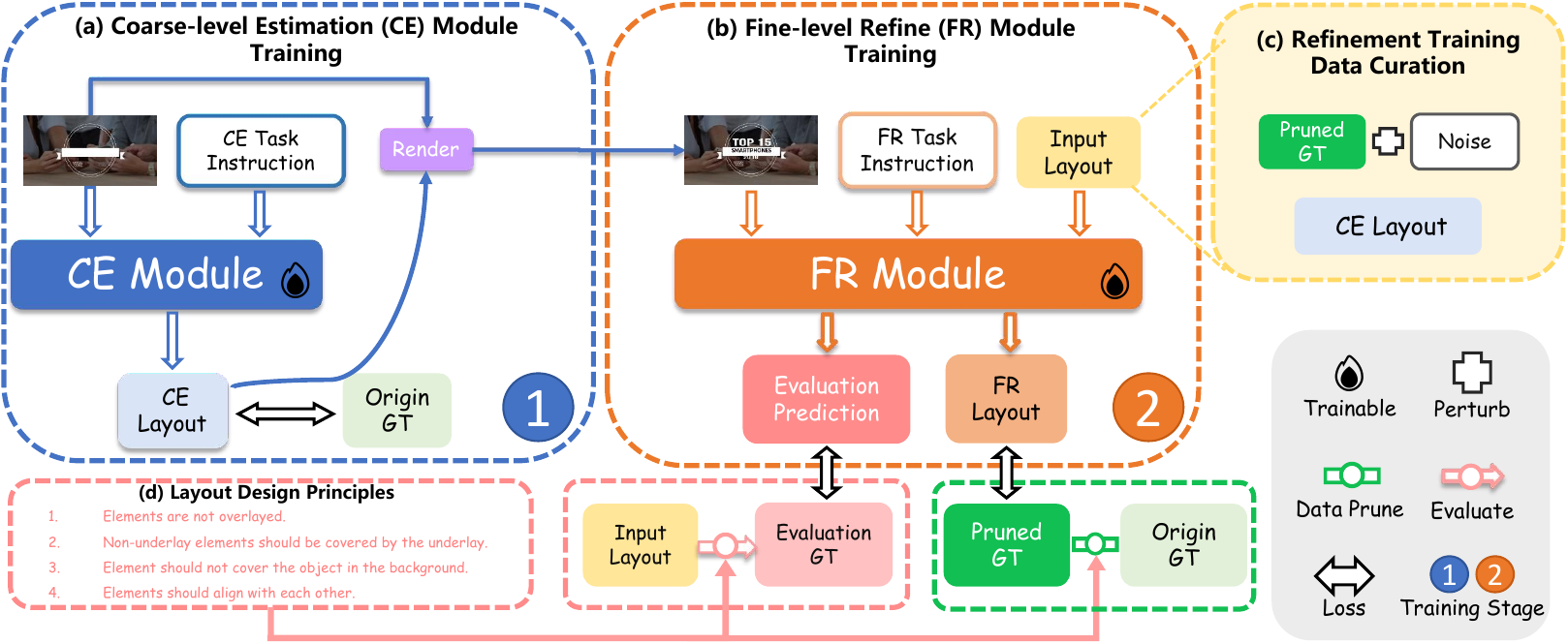}
    \end{center}
           \vspace{-13pt}
    \caption{ \textbf{The training overview of SEGA.} We introduce the training process of the coarse-level estimation model and the fine-level refinement model, as shown in (a) and (b).}
    \vspace{-10pt}
    \label{fig:method}
\end{figure*}
\section{Method}
\label{sec:method}

In this section, we elaborate on our method SEGA as four parts. Specifically, Section \ref{subsec:Overview} introduces the overview of our method and define the symbol usage. Section \ref{subsec:CE-module} and \ref{subsec:FR-module} introduces two main components of our model: Coarse-level Estimation module (CE Module) and Fine-level Refinement module (FR Module), respectively. Section \ref{subsec:Inference} presents the overall inference strategy of our method.

\subsection{Overview}
\label{subsec:Overview}
Aiming to enhance the layout generation ability, our SEGA method is constructed by employing a coarse-to-fine reasoning paradigm. We first utilize the CE module to produce the rough intermediate layout, then leverage the FR module to conduct self-correction based on the former. Since our CE and FR modules are cascaded, the two training datasets feeding into them, $\mathcal{D}_{\textit{CE}}$ and $\mathcal{D}_{\textit{FR}}$, are defined as follows: 
\begin{equation}
\begin{aligned}
\mathcal{D}_{\textit{CE}}  =  & \left \{( (\textit{I}_{i},\textit{T}_{i}^{\textit{CE}}), \textit{L}_{i}^{\textit{GT}} )\right \}_{i=1}^{M} , \\
\mathcal{D}_{\textit{FR}}  = & \left \{((\textit{I}_{j}^{*},\textit{T}_{j}^{\textit{FR}},L_{j}^{\textit{CE}} ), (\textit{E}_{j}^{GT} ,\textit{L}_{j}^{\textit{GT}} ))\right \}_{j=1}^{N} ,
\end{aligned}
\end{equation}
where $\textit{I}$ represents the image, $\textit{I}^{*}$ represents the image with rendered layout, $\textit{T}^{\textit{CE}}$ and $\textit{T}^{\textit{FR}}$ represents task instruction, $\textit{L}^{\textit{GT}}$ represents the ground truth layout, $ \textit{L}^{\textit{Input}} $ represents the layout to be refined, $\textit{E}^{GT}$ represents the evaluation GT for the layout according to design principles. $M$, $N$ indicates the total number of samples in two datasets respectively, and $i$, $j$ represents their position in their respective datasets. Each layout $\textit{L}$ possesses multiple elements $e$, \textit{i.e.}, $\textit{L}  = \left \{ \textit{e}_{1}, \textit{e}_{2}, ..., \textit{e}_{k} \right \}$, where $k$ is the total number of elements in a layout. Following previous works~\cite{zhou2022cgl}, we describe element $\textit{e}$ by its element type $c$, left coordinate $l$, top coordinate $t$, width $w$ and height $h$, \textit{i.e.}, $\textit{e} = (c, l, t, w, h)$. 

\subsection{Coarse-level Estimation Module}
\label{subsec:CE-module}

In the first training stage, the CE module takes a background image and the task instruction as input and output the coarse layout results, which can be formulated as: 
\begin{equation}
\begin{aligned}
L^{\textit{CE}}  = \textit{M}_{\textit{CE}}(I, T^{\textit{CE}}).
\end{aligned}
\label{eq:coarse}
\end{equation}
where the base module $\textit{\textbf{M}}_\textit{CE}$ is built upon the structure of LLaVA-1.5~\cite{liu2024improved}, and we adopt the instruction tuning technique to train our model.


\subsection{Fine-level Refinement Module}
\label{subsec:FR-module}

Given the background image $\textit{I}$ and coarse layout results $\textit{CE}$, inspired by recent research on visual prompt~\cite{yang2023set,cai2023making}, we build an intermediate background image by rendering the elements on initial background image $\textit{I}$ according to $\textit{CE}$, as shown in Figure~\ref{fig:method} (b). This process can be formulated as:   
\begin{equation}
\begin{aligned}
\textit{I}^{*} = Render( I,L^{\textit{FR}} ).
\end{aligned}
\label{eq:render}
\end{equation}

Afterwards, in the second training stage, the FR module takes in intermediate background image $\textit{I}^{*}$, fine-level task instruction $\textit{T}^{\textit{RF}}$ and coarse layout $\textit{L}_{j}^{\textit{CE}}$, then simultaneously predicts the layout evaluation and refined layout results, which can be written as:  
\begin{equation}
\begin{aligned}
E^{\textit{Pred}},L^{\textit{FR}}  = \textbf{\textit{M}}_{\textit{FR}}(I^{*}, T^{\textit{FR}},L^{\textit{\textit{Input}}} ).
\end{aligned}
\end{equation}
where the FR module $\textbf{\textit{M}}^\textit{FR}$ is initialized from the trained CE module $\textit{\textbf{M}}_\textit{CE}$ to get more prior layout knowledge.

To further improve the capacity of model to handle more complex scenes, we proposed two strategies: First, in Section \ref{subsec:Data Curation}, we curate more diverse training data; Second, in Section \ref{sec:prior}, we leverage prior principles to promote the graphic reasoning ability of model on layout designing.

\subsubsection{Refinement Training Data Curation}
\label{subsec:Data Curation}
During the iteration process, the input of the FR module changes from poor to good, which requires our training data to be diverse. And the adjusted layout is expected to be better, which requires high data quality.

We enrich the diversity of input layouts from two perspectives: collecting the output layout of CE module $L^{\textit{CE}}$and creating perturbation layout $L^{\textit{Perturbed}}$, written as 
\begin{align}
L^{\textit{Input}} & = \begin{cases}
L^{\textit{CE}} & p> \epsilon,\\
L^{\textit{Perturbed}}, & p \le \epsilon,
\end{cases}
\end{align}
where $p$ follows a uniform distribution between 0 and 1 and $\epsilon$  represents the threshold.
To avoid data leakage, we can only both train the CE module and infer the CE module to collect $\textit{L}^{\textit{CE}}$ in the same training splits. However, CE module has learned the $\textit{L}^{\textit{GT}}$ in its training stage, which leads to the fact that the layouts we collected directly are much better than the layouts to be refined in the test stage. To avoid such a domain gap, we pick a CE checkpoint with fewer training epochs to collect the layout. As for the $\textit{L}^{\textit{Perturbed}}$, we randomly select an element in the $\textit{L}^{\textit{GT}}$ and perturb it, and the detailed pseudo code of the perturbing process can be found in Section 3.2 of our appendix.

To ensure the high quality of training data, we use the design principles proposed in Section~\ref{sec:prior} to prune the data, which removes layouts that do not meet the prior from the original training set (see details in our appendix).

\begin{table*}[t]
\renewcommand\arraystretch{1.1}	
        \begin{center}
        \resizebox{1\linewidth}{!}{

        \begin{tabular}{ccccccccccccc}
        \toprule[0.1pt]
        \toprule[0.1pt]
				Dataset & Layout Amount & No Artifact & Fine-grained Attributes  & Structured & Clustered & Average Elements & Variance Elements & Text &  \\      
				\midrule			
				PKU  & 	9,974 &  \XSolidBrush	 & \XSolidBrush	  & 	\XSolidBrush & \XSolidBrush	 & 	4.53 & 	3.39 &  \XSolidBrush  &     \\ 
				
				CGL  &	60,548  &	\XSolidBrush  &	\XSolidBrush  &	\XSolidBrush  &	 \XSolidBrush  &	4.80  &	3.84  &  \Checkmark  &\\ 
				
				Crello  &	23,182  & \Checkmark  &	\Checkmark  &	\XSolidBrush  &	 \XSolidBrush &	4.29 &	10.23  &  \Checkmark   &             \\

				\midrule
				
				\textbf{GenPoster-100K}  &	105,456  &	\Checkmark  &	\Checkmark  &	\Checkmark  & \Checkmark &	5.85  &	34.90    &   \Checkmark  &\\				
		\bottomrule[0.1pt]
            \bottomrule[0.1pt]
            \end{tabular}
            }
            \end{center}
        \vspace{-13pt}
            \caption{Statistics comparison between different graphic layout datasets}
            \vspace{-13pt}
\label{tab:datasetcomparison}
\end{table*}

\subsubsection{Layout Evaluation with Design Principles}
\label{sec:prior}

Typically, for human being, optimal decisions are made only when we have a comprehensive and accurate understanding of one task. Inspired by this, we enable the model to grasp human design principles for graphic layout planning through instruction tuning. Based on these principles, we evaluate existing layout results; if a layout is deemed irrational, iterative optimization can be applied to further refine it. 
Concretely, we briefly summarize the principles for layout generation as four points: (\textit{\romannumeral 1}) \textit{Elements are not overlaid.}
(\textit{\romannumeral 2}) \textit{Non-underlay elements should be covered by the underlay.}
(\textit{\romannumeral 3}) \textit{Elements should not cover the object in the background.}
(\textit{\romannumeral 4}) \textit{Elements should align with each other. The corresponding evaluation is defined based on the rule-based metric calculation.} 
Specifically, if the input layout violates certain design principle, the evaluation results will reveal it. For example, if principle 1) is violated, our evaluation will be written as: \texttt{there is element overlap in the current poster}; If the input layout meets all principles, it is written as: \texttt{current layout is fine}. Note that with aid of the auto-regressive learning manner, the evaluation results could directly affect the predicting procedure of output layout results, namely Evaluation Chain-of-Thought (ECoT) in our method.

\subsection{The Iterative Inference of SEGA}
\label{subsec:Inference}
In the inference stage, the user inputs a background image $I$ and a conditional constraint $C$, and we first integrate them with task description $D$ to form the input for CE module, i.e., $ ( I, \mathcal{D}_{\textit{CE}}, C )$. Then, we get the intermediate layout $L^{\textit{CE}}$ from Eq.~\ref{eq:coarse}, which is regarded as $L^{\textit{Input}}_{\textit{Pred}_{0}}$. With Eq.~\ref{eq:render}, we can render the layout onto the background image $I$ and get the input $( I^{*},T_{\textit{FR}}, L_{\textit{Pred}_{k}}^{\textit{Input}} )$ for the FR module, where $k$ represents the number of iteration. The iterative refinement can be written as: 
\begin{equation}
\begin{aligned}
E^{\textit{Pred}}_{k},L_{\textit{Pred}_{k+1}}^{\textit{FR}}  = M_{\textit{FR}}( I^{*} ,T_{\textit{FR}} ,L_{\textit{Pred}_{k}}^{\textit{FR}} ).
\end{aligned}
\end{equation}
 We regard $L_{\textit{Pred}_{k+1}}^{\textit{FR}}$ as final layout output after appropriate iteration rounds, which is discussed in the Section~\ref{s:Ablation}.

\subsection{Discussion}

Here we compare our method with related studies that adopt similar refinement strategies to enhance layout generation capabilities. Specifically, Lee et al.~\cite{lee2020neural} employ a three-module pipeline for sequential graph prediction, layout generation, and iterative refinement via fine-tuning. Meanwhile, Iwai et al.~\cite{iwai2024layout} introduce a "Corrector" mechanism tailored for layout sticking tasks. 

Our approach diverges from these works in two key aspects: First, the task settings are different. We address content-aware layout generation, whereas prior methods focus on content-unaware scenarios. Second, fundamental architectures are different. Unlike graph neural network~\cite{lee2020neural} trained from scratch or diffusion-based element resetting~\cite{iwai2024layout}, our method is constructed based on Large Language Models (LLMs). To fully invoke the language comprehension ability of LLM, with the auto-regressive paradigm learning, we exploit the evaluation results on previous layouts as feedback to enlighten current layout design, thereby elevating layout generation fidelity.

\begin{table*}[t]
\small
\renewcommand\arraystretch{1.06}	
\begin{center}
\resizebox{0.95\linewidth}{!}{
\begin{tabular}{ccccccccccccc}
\toprule[0.1pt]
\toprule[0.1pt]
\multirow{3}{*}{Method} & \multicolumn{6}{c}{PKU} & \multicolumn{6}{c}{CGL} \\ 
\cmidrule(lr){2-7}\cmidrule(lr){8-13}
& \multicolumn{4}{c}{Graphic} & \multicolumn{2}{c}{Content} & \multicolumn{4}{c}{Graphic} & \multicolumn{2}{c}{Content} \\ 
\cmidrule(lr){2-5}\cmidrule(lr){6-7}\cmidrule(lr){8-11}\cmidrule(lr){12-13}
& \textit{Ali} $\downarrow$ & \textit{Ove} $\downarrow$ & \textit{Und\_l} $\uparrow$ & \textit{Und\_s} $\uparrow$ & \textit{Read} $\downarrow$ & \textit{Occ} $\downarrow$ 
& \textit{Ali} $\downarrow$ & \textit{Ove} $\downarrow$ & \textit{Und\_l} $\uparrow$ & \textit{Und\_s} $\uparrow$ & \textit{Read} $\downarrow$ & \textit{Occ} $\downarrow$ \\ 
\hline
\textbf{Non-LLM Based} & & & & & & & & & & & & \\ 
CGL-GAN~\cite{zhou2022cgl}~(IJCAI, 2022) 
& - & 0.0380 & - & 0.4800 & 0.0158 & 0.1320 
& - & 0.0470 & - & 0.6500 & 0.0213 & 0.1400 \\ 
LayoutDM~\cite{inoue2023layoutdm}~(CVPR, 2023) 
& - & 0.1720 & - & 0.4600 & 0.0201 & 0.1520 
& - & 0.0260 & - & 0.7900 & 0.0192 & 0.1270 \\ 
RALF~\cite{horita2024retrieval}~(CVPR, 2024) 
& \textbf{0.0031} & 0.0095 & 0.9686 & 0.8981 & \textbf{0.0138} & \textbf{0.1243} 
& 0.0023 & 0.0059 & 0.9858 & 0.9652 & \textbf{0.0180} & \textbf{0.1263} \\ 
\hline
\textbf{LLM Based} & & & & & & & & & & & & \\ 
PosterLlama~\cite{seol2024posterllama}~(ECCV, 2024) 
& 0.0036 & 0.0080 & 0.9874 & 0.9497 & 0.0170 & 0.1380 
& 0.0022 & 0.0042 & 0.9823 & 0.9463 & 0.0294 & 0.2453 \\ 
SEGA w/o FR 7B 
& 0.0037 & 0.0052 & 0.9873 & 0.9471 & 0.0150 & 0.1336 
& 0.0023 & 0.0032 & 0.9817 & 0.9522 & 0.0298 & 0.2442 \\ 
SEGA w/o FR (Ens-2) 7B 
& 0.0035 & 0.0041 & 0.9892 & 0.9673 & 0.0144 & 0.1305 
& 0.0021 & 0.0024 & 0.9879 & 0.9657 & 0.0296 & 0.2438 \\ 
SEGA (7B) 
& 0.0035 & \textbf{0.0033} & \textbf{0.9897} & \textbf{0.9731} & 0.0142 & 0.1286 
& \textbf{0.0020} & \textbf{0.0017} & \textbf{0.9913} & \textbf{0.9782} & 0.0294 & 0.2430 \\ 
\textbf{GT} 
& 0.0036 & 0.0009 & 0.9950 & 0.9444 & 0.0119 & 0.1185 
& 0.0023 & 0.0003 & 0.9937 & 0.9402 & 0.0296 & 0.2390 \\
\bottomrule[0.1pt]
\bottomrule[0.1pt]
\end{tabular}
}
\end{center}
\vspace{-15pt}
\caption{\textbf{Experimental results on the PKU and CGL dataset.} For better demonstration, we \textbf{bold} the best values.}
\label{tab:main_result_pkucgl}
\vspace{-5pt}
\end{table*}

\begin{table*}[t]
\normalsize
\renewcommand\arraystretch{1.06}
\begin{center}
\resizebox{0.95\linewidth}{!}{
\begin{tabular}{ccccccc|ccccc|c}
\toprule[0.1pt]
\toprule[0.1pt]
\multirow{3}{*}{Method}                                                        & \multicolumn{6}{c|}{Rule-based Metrics} & \multicolumn{5}{c|}{Aesthetic Scores} & \multirow{3}{*}{\makecell{Inference\\Time~(s)}} \\ \cline{2-12}
                                                                               & \multicolumn{4}{c}{Graphic} & \multicolumn{2}{c|}{Content} & \multirow{2}{*}{$\textit{S}_{\text{DL}} \uparrow$} & \multirow{2}{*}{$\textit{S}_{\text{QL}} \uparrow$} & \multirow{2}{*}{$\textit{S}_{\text{TV}} \uparrow$} & \multirow{2}{*}{$\textit{S}_{\text{IO}} \uparrow$} & \multirow{2}{*}{$\textit{S}_{\text{Mean}} \uparrow$} & \\ \cline{2-7}
                                                                               & $\textit{Ali} \downarrow$ & $\textit{Ove} \downarrow$ & $\textit{Und\_l} \uparrow$ & $\textit{Und\_s} \uparrow$ & $\textit{Read} \downarrow$ & $\textit{Occ} \downarrow$ & & & & & & \\ \hline
\textit{\textbf{Non-LLM Based}}                                                & & & & & & & & & & & & \\
FlexDM~\cite{inoue2023towards}~(CVPR, 2023)         & 0.0122 & 0.1139 & 0.6889 & 0.5034 & 0.0516 & 0.4850 & 4.563 & 5.126 & 4.873 & 5.239 & 4.950 & 2.03 \\ \cline{1-12}
\textit{\textbf{LLM Based}}                                                    & & & & & & & & & & & & \\
PosterLlama~\cite{seol2024posterllama}~(ECCV, 2024) & 0.0099 & 0.0238 & 0.9204 & 0.7378 & 0.0395 & 0.4041 & 5.292 & 5.796 & 5.263 & 5.819 & 5.542 & 7.33 \\
SEGA w/o FR 7B           & 0.0102 & 0.0121 & 0.8206 & 0.6698 & 0.0304 & 0.4002 & 5.553 & 6.332 & 5.693 & 5.448 & 5.756 & 8.76 \\
SEGA w/o FR (Ens-2) 7B & 0.0100 & 0.0093 & 0.8501 & 0.7202 & 0.0285 & 0.3957 & 5.642 & 6.418 & 5.811 & 5.529 & 5.850 & 8.76$\times$2 \\
SEGA 7B                        & \textbf{0.0086} & 0.0040 & 0.9337 & 0.8978 & 0.0282 & 0.3964 & 5.792 & 6.411 & 5.824 & 5.708 & 5.941 & 8.76+7.77 \\
SEGA w/o FR 13B          & 0.0102 & 0.0093 & 0.8485 & 0.7315 & 0.0271 & 0.3948 & 5.923 & 6.624 & 6.253 & 5.991 & 6.197 & 10.28 \\
SEGA w/o FR (Ens-2) 13B & 0.0097 & 0.0075 & 0.8715 & 0.7715 & \textbf{0.0260} & \textbf{0.3874} & 6.128 & 6.652 & 6.058 & 5.822 & 6.165 & 10.28$\times$2 \\
SEGA 13B                        & 0.0095 & \textbf{0.0025} & \textbf{0.9541} & \textbf{0.9270} & \textbf{0.0260} & 0.3907 & \textbf{6.149} & \textbf{6.745} & \textbf{6.348} & \textbf{6.038} & \textbf{6.320} & 10.28+10.01 \\
GT                              & 0.0100 & 0.0116 & 0.9643 & 0.8187 & 0.0259 & 0.3797 & 6.882 & 7.543 & 6.863 & 6.025 & 6.828 & - \\ 
\bottomrule[0.1pt]
\bottomrule[0.1pt]
\end{tabular}
}
\end{center}
\vspace{-15pt}
\caption{\textbf{Experimental results on the Crello dataset.} We train SEGA on the base of 7B and 13B respectively and compare it with a strong baseline with Best of $k$ (Ens-$k$) strategy.}
\label{tab:main_result_crello}
\vspace{-0.7cm}
\end{table*}

\section{Our Dataset: GenPoster-100K}
To facilitate further development in the field of layout generation, we present a new GenPoster-100K dataset, which consists of 105,456 posters and their corresponding hierarchical metadata that includes images of individual elements. To better understand the characteristics of our dataset, we compare it with other publicly available datasets from multiple perspectives, such as data volume, the number and variance of layout elements, and the richness of layout element attribute values, as shown in Table~\ref{tab:datasetcomparison}. Thanks to the diverse and rich design patterns, the model pre-trained on our dataset is able to exhibit satisfactory cross-dataset generalization capabilities (Section \ref{section:more-exp}) and holds the potential to advance numerous related tasks. Due to limited space, the details of GenPoster-100K dataset and more related experiments are given in the appendix.

\section{Experiment}

\subsection{Setup}

\noindent \textbf{--Dataset}
We utilize three publicly available datasets: PKU~\cite{hsu2023posterlayout}, CGL~\cite{zhou2022cgl}, and Crello~\cite{yamaguchi2021canvasvae}. PKU and CGL datasets primarily focus on e-commerce domain, while the Crello dataset covers a wider variety of more complex and diverse poster themes. As no type information of layout elements is given, we use the Yi-34B LLM~\cite{young2024yi} to classify all input text into predefined categories and reorganize the data with a similar format of PKU and CGL, as detailed in Section 4.2 of our appendix.

\noindent \textbf{--Metrics}
Following previous works~\cite{seol2024posterllama, hsu2023posterlayout, li2020attribute}, we use three kinds of metrics to evaluate our method: \textit{Graphic metrics}, \textit{Content metrics}, and \textit{Aesthetic scores}. \textit{Graphic metrics} assess the coherence among predicted elements, including \texttt{Alignment} ($\mathit{Ali}~\mathit{\downarrow}$), \texttt{Overlay} ($\mathit{Ove}~\mathit{\downarrow}$), \texttt{Underlay effectiveness loose} ($\mathit{Und\_l}~\mathit{\uparrow}$), and \texttt{Underlay effectiveness strict} ($\mathit{Und\_s}~\mathit{\uparrow}$). \textit{Content metrics} evaluate the harmony between the predicted elements and the background image, including \texttt{Readability score} ($\mathit{Rea}~\mathit{\downarrow}$) and \texttt{Occlusion} ($\mathit{Occ}~\mathit{\downarrow}$). Moreover, unlike two kinds of rule-based criteria above, we follow \cite{jia2023cole,cheng2025graphic} to utilize GPT-4V~\cite{hurst2024gpt} to comprehensively evaluate the quality of final rendered images as \textit{Aesthetic scores} from four aspects ($\textit{S}_{DL}~\mathit{\uparrow}$, $\textit{S}_{QL}~\mathit{\uparrow}$, $\textit{S}_{TV}~\mathit{\uparrow}$, $\textit{S}_{IO}~\mathit{\uparrow}$). More detailed description of all metrics are illustrated in Section 4.3 of our appendix.

 \begin{figure}[t]
    \begin{center}
        \includegraphics[width=\linewidth]{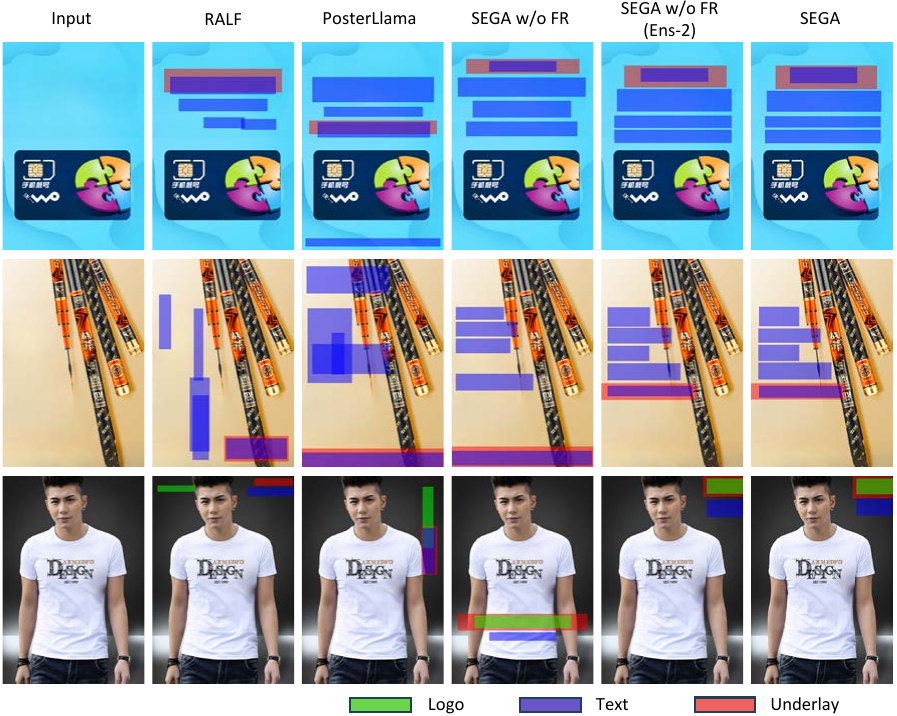}
    \end{center}
           \vspace{-15pt}
    \caption{Qualitative visualization comparison in PKU dataset.} %

    \label{fig:pku_cgl}
    \vspace{-16pt}
\end{figure}

 \begin{figure*}[t]
    \begin{center}
        \includegraphics[width=0.93\linewidth]{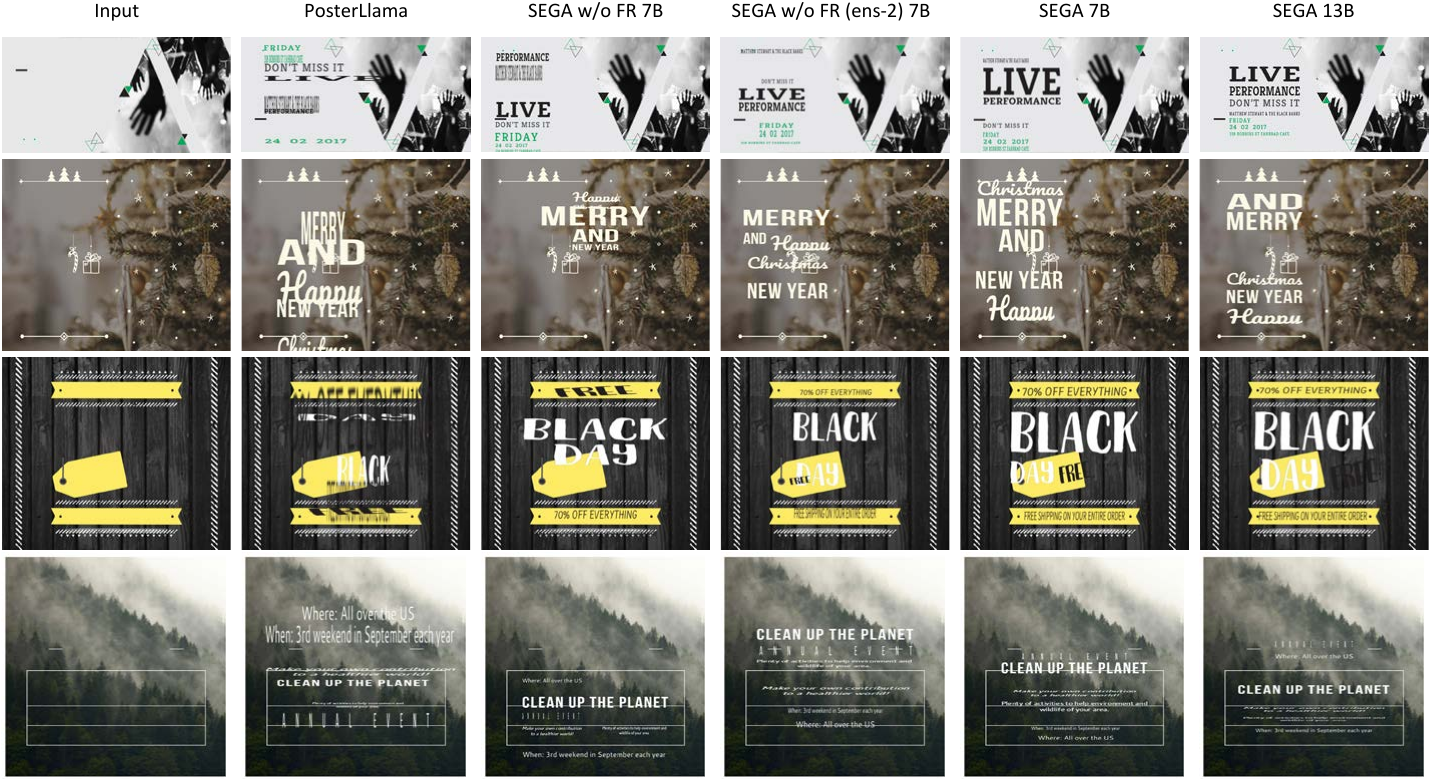}
    \end{center}
           \vspace{-0.25in}
    \caption{ \textbf{Qualitative visualization comparison in the Crello dataset.} Since this dataset is layer-independent, we present the rendered results of the layout predicted by the model.} %

    \label{fig:crello}
    \vspace{-0.23in}
\end{figure*}

\noindent \textbf{--Methods in Experiments}
In our experiments, we compare our approach with five other SoTA methods. CGL-GAN~\cite{zhou2022cgl}, FlexDM~\cite{inoue2023towards}, and RALF~\cite{horita2024retrieval} are non-LLM-based methods. PosterLlama~\cite{seol2024posterllama} is an LLM-based layout generation method that formulate layout information into HTML code. Our baseline is SEGA w/o FR, which only comprises the CE module. SEGA w/o FR (Ens-$k$) represents the best results ensembled from model output with $k$ repeated times. SEGA is our full model being composed of both CE and FR module. 

\noindent \textbf{--Implementation Details}
We employ LLaVa-1.5~\cite{liu2024improved} (7B, 13B) as our base model, meanwhile reuse its most hyperparameter setup. Since no experiments on Crello dataset are conducted in PosterLlama~\cite{seol2024posterllama}, we use its source code to replicate the corresponding results. For the results, we generate layouts on three independent trials and report the average of the metrics for fairness. More details can be found in Section 4.1 of our appendix.

\subsection{Comparison with SoTA Methods}
Here, we present experimental results under constraint-based setting \cite{zhou2022cgl,horita2024retrieval} that provides fixed layout element types. The experiments under unconstrained setting~\cite{jiang2023layoutformer} that generates layout without any user requirements are given in Section 4.5 of our appendix for restricted space.

\noindent \textbf{--Quantitative Results.}~
Table~\ref{tab:main_result_pkucgl} and Table~\ref{tab:main_result_crello} present the quantitative results of the constraint-based task. First, it can be seen that our baseline SEGA w/o FR has obvious advantage over the previous methods on most metrics, which confirms the strong potential of MLLM for layout generation. In comparison to the voting based method SEGA w/o FR (Ens-$2$), our SEGA achieves superior performance on 11 of the 12 metrics on CGL and PKU datasets. Overall, our SEGA significantly outperforms existing methods on all three datasets. As for the weakness of Content Metics on the CGL dataset, we speculate that the potential reason is that the background images obtained through inpainting may degrade the representational capability of the visual encoder in MLLMs. However, SEGA is still superior to the LLM-based method PosterLlama. In addition, our method is leading on both base models of 7B and 13B, which demonstrates the scalability of our framework. 


\noindent \textbf{--Qualitative Results.}~
We present qualitative visualization results on the PKU dataset in Figure~\ref{fig:pku_cgl} and leave that on the CGL dataset in Section 4.8 of the Appendix, showing our SEGA generates non-overlapping, well-designed layouts. For the Crello dataset, we combine the elements from separate layers and render them on the background image. From Figure~\ref{fig:crello}, the results highlight our advantage on more complex layout scenes. 



\begin{table*}[t]
\small
\renewcommand\arraystretch{1.06}
 \begin{center}
    \resizebox{0.90\linewidth}{!}{

\begin{tabular}{cccccccccccc}
\toprule[0.1pt]
\toprule[0.1pt]
\multirow{2}{*}{Config} & \multicolumn{3}{c}{Data}                                                          & \multirow{2}{*}{ECoT} & \multirow{2}{*}{Visual Prompt} & \multicolumn{4}{c}{Graphic}                                                               & \multicolumn{2}{c}{Content} \\ \cmidrule(lr){2-4} \cmidrule(lr){7-10} \cmidrule(lr){11-12}
                        & \multicolumn{1}{c}{Data Collection } & \multicolumn{1}{c}{Pertubed Data} & Prune Data   &                      &                       & \multicolumn{1}{c}{\textit{Ali} $\downarrow$}   & \multicolumn{1}{c}{\textit{Ove} $\downarrow$}    & \multicolumn{1}{c}{\textit{Und\_l} $\uparrow$}  & \multicolumn{1}{c}{\textit{Und\_s} $\uparrow$}  & \multicolumn{1}{c}{\textit{Read} $\downarrow$}   & \multicolumn{1}{c}{\textit{Occ} $\downarrow$}    \\ \hline
1                       & \multicolumn{1}{c}{}             & \multicolumn{1}{c}{}              &              &                      &                                & \multicolumn{1}{c}{0.0102} & \multicolumn{1}{c}{0.0093} & \multicolumn{1}{c}{0.8485} & 0.7315 & \multicolumn{1}{c}{0.0271} & 0.3948 \\ 
2                       & \multicolumn{1}{c}{$\checkmark$} & \multicolumn{1}{c}{}              &              &                      &                                & \multicolumn{1}{c}{0.0098} & \multicolumn{1}{c}{0.0125} & \multicolumn{1}{c}{0.9336} & 0.8076  & \multicolumn{1}{c}{0.0299} & 0.3922 \\ 
3                       & \multicolumn{1}{c}{$\checkmark$} & \multicolumn{1}{c}{$\checkmark$}  &              &                      &                                & \multicolumn{1}{c}{0.0093} & \multicolumn{1}{c}{0.0127} & \multicolumn{1}{c}{0.9440} & 0.8244  & \multicolumn{1}{c}{0.0274} & 0.3897 \\ 
4                       & \multicolumn{1}{c}{$\checkmark$} & \multicolumn{1}{c}{$\checkmark$}  & $\checkmark$ &                      &                                & \multicolumn{1}{c}{0.0084} & \multicolumn{1}{c}{0.0062} & \multicolumn{1}{c}{0.9394} & 0.7948  & \multicolumn{1}{c}{0.0276} & 0.3876 \\ 
5                      & \multicolumn{1}{c}{$\checkmark$} & \multicolumn{1}{c}{$\checkmark$}  & $\checkmark$ & $\checkmark$         &                                & \multicolumn{1}{c}{\textbf{0.0081}} & \multicolumn{1}{c}{0.0056} & \multicolumn{1}{c}{0.9401} & 0.8123  & \multicolumn{1}{c}{0.0263} & \textbf{0.3835} \\ 
6                       & \multicolumn{1}{c}{$\checkmark$} & \multicolumn{1}{c}{$\checkmark$}  & $\checkmark$ & $\checkmark$         & $\checkmark$                   & \multicolumn{1}{c}{0.0095} & \multicolumn{1}{c}{\textbf{0.0025}} & \multicolumn{1}{c}{\textbf{0.9541}} & \textbf{0.9270}  & \multicolumn{1}{c}{\textbf{0.0260}} & 0.3907 \\ 
\bottomrule[0.1pt]
\bottomrule[0.1pt]
\end{tabular}
}
\end{center}
        \vspace{-0.20in}
\caption{The impact of different components, including training data curation, ECoT for inference and input visual prompt.} 
    \label{tab:Ablation}
        \vspace{-0.10in}
\end{table*}

\begin{table*}[t]
\renewcommand\arraystretch{1.06}
\begin{center}
\resizebox{0.9\linewidth}{!}{
\begin{tabular}{cccccccc|ccccc}
\toprule[0.1pt]
\toprule[0.1pt]
                           & \multirow{3}{*}{Method}     & \multicolumn{6}{c|}{Rule-based Metrics} & \multicolumn{5}{c}{Aesthetic Scores} \\ \cline{3-13} 
                           &                             & \multicolumn{4}{c}{Graphic} & \multicolumn{2}{c|}{Content} & \multirow{2}{*}{$\textit{S}_{\text{DL}} \uparrow$} & \multirow{2}{*}{$\textit{S}_{\text{QL}} \uparrow$} & \multirow{2}{*}{$\textit{S}_{\text{TV}} \uparrow$} & \multirow{2}{*}{$\textit{S}_{\text{IO}} \uparrow$} & \multirow{2}{*}{$\textit{S}_{\text{Mean}} \uparrow$} \\ \cline{3-8}
                           &                             & $\textit{Ali} \downarrow$ & $\textit{Ove} \downarrow$ & $\textit{Und\_l} \uparrow$ & $\textit{Und\_s} \uparrow$ & $\textit{Read} \downarrow$ & $\textit{Occ} \downarrow$ & & & & & \\ \hline
\multirow{2}{*}{Zero-shot} & CGL Pre-train               & \textbf{0.0068} & \textbf{0.0014} & 0.5564 & 0.2999 & 0.0419 & 0.4002 & 5.132 & 5.922 & 5.258 & 5.390 & 5.425 \\
                           & GenPoster-100K Pre-train    & 0.0103 & 0.0125 & 0.7711 & 0.6673 & 0.0247 & 0.4000 & 5.844 & 6.604 & 5.97 & 5.592 & 6.002 \\ \hline
\multirow{6}{*}{Finetune}  & SEGA w/o FR                 & 0.0102 & 0.0121 & 0.8206 & 0.6698 & 0.0304 & 0.4002 & 5.553 & 6.332 & 5.693 & 5.448 & 5.756 \\
                           & SEGA w/o FR $*$ Crello      & 0.0116 & 0.0138 & 0.8146 & 0.6713 & 0.0321 & 0.4098 & 5.541 & 6.319 & 5.657 & 5.430 & 5.736 \\
                           & SEGA w/o FR $*$             & 0.0073 & 0.0093 & 0.8350 & 0.7515 & 0.0237 & \textbf{0.3787} & 5.868 & 6.668 & 6.038 & 5.636 & 6.052 \\
                           & SEGA                        & 0.0086 & 0.0040 & 0.9337 & 0.8978 & 0.0282 & 0.3964 & 5.792 & 6.411 & 5.824 & 5.708 & 5.941 \\
                           & SEGA $*$ Crello             & 0.0092 & 0.0052 & 0.9249 & 0.8831 & 0.0291 & 0.3977 & 5.786 & 6.403 & 5.828 & 5.701 & 5.930 \\
                           & SEGA$*$                     & 0.0077 & 0.0038 & \textbf{0.9580} & \textbf{0.9298} & \textbf{0.0221} & 0.3844 & \textbf{6.016} & \textbf{6.752} & \textbf{6.070} & \textbf{5.710} & \textbf{6.137} \\ 
\bottomrule[0.1pt]
\bottomrule[0.1pt] 
\end{tabular}
}
\end{center}
\vspace{-15pt}
\caption{The impact of pre-training on the GenPoster-100K Dataset. The results show that pre-training on the GenPoster-100K dataset is more effective than the CGL dataset and bring performance gains. The CGL and GenPoster-100K pre-train represent the results of direct testing on Crello after pre-training on those two datasets. $*$ and $*$ Crello represent the model pretrained on the GenPoster-100K and Crello dataset, respectively.}
\label{tab:pretrain}
\vspace{-15pt}
\end{table*}

\subsection{Ablation Study and Analysis}
\label{s:Ablation}

In this section, unless otherwise specified, all the experiments are carried out on the most challenging Crello~\cite{yamaguchi2021canvasvae} dataset and based on LLaVa-1.5~\cite{liu2024improved} 13B model. 

First, we conduct a series of ablation experiments to testify the influence of different components in our SEGA module. Due to limited space, the impact of three crafts, including training data collection, ECoT, and visual prompt, more ablation studies can be found in Section 4.6 of our appendix. As shown in Table \ref{tab:Ablation}, the strategy of our training data collection (Config 4) brings corresponding improvements over the original method (Config 1), indicating the importance of diversity of training data. In addition, observing Config 5 and Config 6, the incorporation of ECoT and visual prompt can deliver clear effectiveness gains, respectively. 

\begin{figure}[t]
    \begin{center}
        \includegraphics[width=\linewidth]{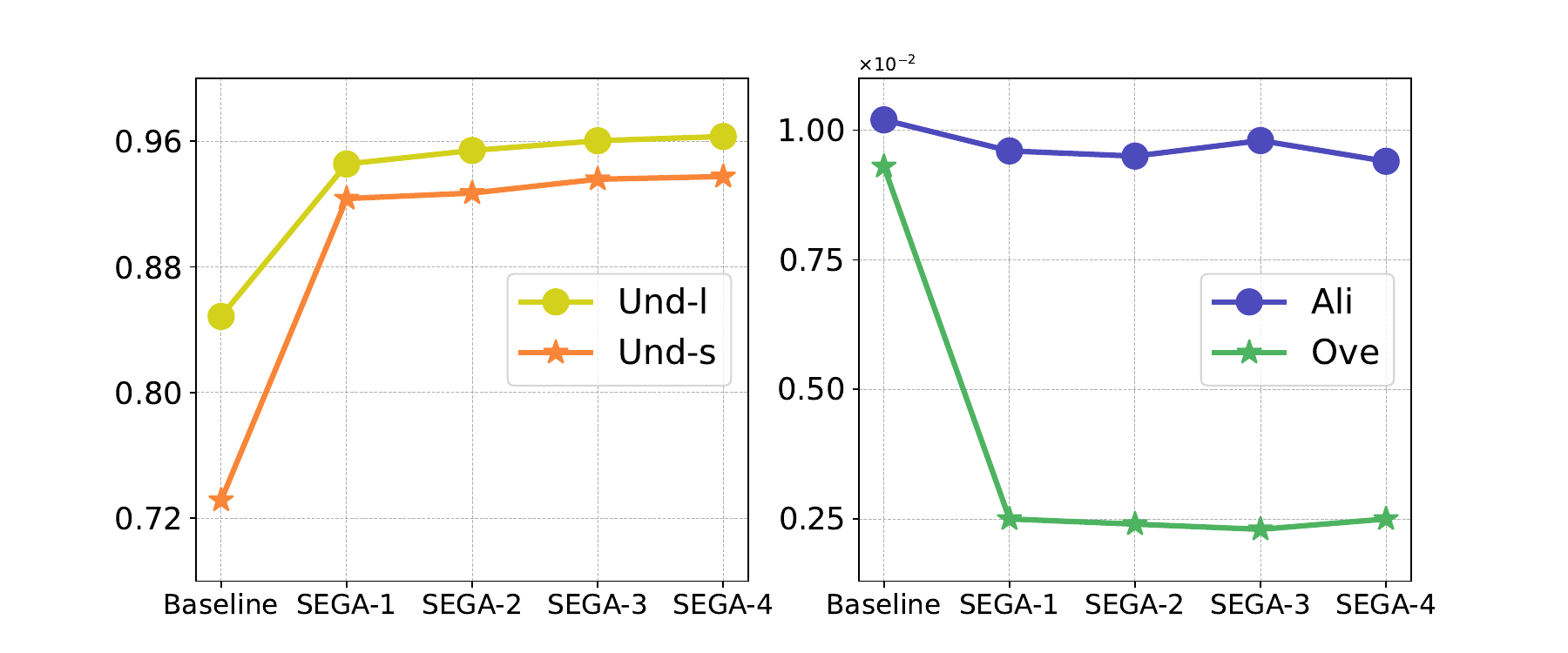}
    \end{center}
           \vspace{-15pt}
    \caption{The impact of iterations of SEGA. As iteration rounds increases, the model performance gradually improves and tends towards saturation. The two metrics in the left figure are better with higher values, while it is the opposite in the right figure.
 } %
    \label{fig:iteration}
    \vspace{-0.6cm}
\end{figure}

Then, the experiments are made to display the impact of iteration refinement rounds during inference stage. As depicted in Figure~\ref{fig:iteration}, one-time iteration leads to the substantial performance gain and more rounds of iteration yields relatively restricted improvement. In light of the trade-off between computational cost and performance amelioration, one-time iteration is chosen as the default setting for our SEGA model.

\begin{table}[t]
\renewcommand\arraystretch{1.08}	
\normalsize 
 \begin{center}
        \resizebox{\linewidth}{!}{

\begin{tabular}{ccccccc}
\toprule[0.1pt]
\toprule[0.1pt]
\multirow{2}{*}{Setting}  & \multicolumn{4}{c}{Graphic}                                                                                                                                                                                                             & \multicolumn{2}{c}{Content}                                                                                        \\ \cmidrule(lr){2-5}\cmidrule(lr){6-7}
                         & \textit{Ali} $\downarrow$ & \textit{Ove} $\downarrow$ & \textit{Und\_l} $\uparrow$ & \textit{Und\_s} $\uparrow$ & \textit{Read} $\downarrow$ & \textit{Occ} $\downarrow$ \\ \hline
PosterLlama& 0.0099                                                  & 0.0238                                                 & 0.9204                                                   & 0.7378                                                   & 0.0395                                                   & 0.4041                                                  \\
PosterLlama + FR & 0.0093                                                  & \textbf{0.0162}                                                  & 0.9314                                                   & 0.8579                                                   & 0.0340                                                   & 0.4108 \\ 
\hline

SEGA w/o FR 7B           & 0.0102                        & 0.0121                        & 0.8206                         & 0.6698                         & 0.0304                         & 0.4002  \\

CE + GPT refine 7B             & 0.0103                        & 0.0281                        & 0.8465                         & 0.7438                         & 0.0331                         & \textbf{0.3953}   \\ 
SEGA  7B                        & \textbf{0.0086}                        & 0.0040                        & \textbf{0.9337}                         & \textbf{0.8978}                         & \textbf{0.0282}                         & 0.3964    \\

\bottomrule[0.1pt]
\bottomrule[0.1pt]
\end{tabular}
}
\end{center}
        \vspace{-0.2in}
\caption{The experimental results of generalization ability of our SEGA framework.}
    \label{tab:general_posterllama}
  \vspace{-16pt}
\end{table}


\subsection{More Comparison Experiments}
\label{section:more-exp}
\subsubsection{Framework Generalization of SEGA}
We explore the generalization ability of the SEGA framework by replacing the CE module with PosterLlama and the FR module with GPT-4o~\cite{hurst2024gpt}. As shown in Table~\ref{tab:general_posterllama}, our framework is still effective but lags behind our SEGA.

\subsubsection{Study on GenPoster-100K as Pre-training Dataset}
In downstream layout generation tasks, different sub-scenarios have different data distributions, such as posters, web pages, etc. In order to reduce data collection costs and enhance the generalization, a common idea is to pre-train on a large layout dataset to obtain a strong base model. Here, we discuss the benefit of GenPoster-100K as a pre-training dataset. All pre-training method is SFT and 
 we test all models in the Crello dataset. The results are shown in Table~\ref{tab:pretrain}. There are three points worth noting: First, Our GenPoster-100K is better than the existing dataset CGL in pretraining for Crello. Second, the pre-trained model is better than the specialized model trained on Crello. Thirdly, our method can complement pre-training. Moreover, the pre-training on the GenPoster-100K dataset is also useful for the PKU and CGL dataset, and the corresponding results can be found in Section 4.7 of our appendix.

\subsubsection{Results in Different Scenes}
We list experiments on both difficult scenes (The number of items to be planned is greater than 8) and normal scenes. From Table~\ref{tab:diffcult}, our method can get significant performance improvement under difficult scenes.

\begin{table}[]
\renewcommand\arraystretch{1.04}
\footnotesize
 \begin{center}
        \resizebox{1.0\linewidth}{!}{
\begin{tabular}{c|cccccc}
\bottomrule[0.1pt]
\bottomrule[0.1pt]
        \multirow{2}{*}{Methods}       & \textit{Ali} $\downarrow$       & \textit{Ove} $\downarrow$                                            & \textit{Und\_l} $\uparrow$        & \textit{Und\_s} $\uparrow$     & \textit{Read}  $\downarrow$        & \multicolumn{1}{c}{\textit{Occ} $\downarrow$}                                                                   \\ \cline{2-7} 
   
             & \multicolumn{6}{c}{Elements \textless{}= 8}                      
             \\ \cline{1-7} 
SEGA w/o FR & 0.0110                                         & 0.0103                                         & 0.8396                                        & 0.7049                                        & 0.0291                                       & \multicolumn{1}{c}{0.4008}                                       \\
\multirow{2}{*}{SEGA}      & 0.0093   & 0.0038  & 0.9668 & 0.9266 & 0.0274  & \multicolumn{1}{c}{0.3974} \\
& \color{green}+22\% & {\color{green}+171\%}  & {\color{green}+15\%} & {\color{green}+31\%} & {\color{green}+6\%} &   {\color{green}+1\%} \\
\cline{1-7} 
                   & \multicolumn{6}{c}{Elements \textgreater 8}                                                                                                                                                                                                                                                         \\ \cline{2-7} 
SEGA w/o FR & 0.0007                                         & 0.0114                                         & 0.7226                                        & 0.5430                                        & 0.0365                                       & 0.3868                                                            \\
\multirow{2}{*}{SEGA}  & 0.0003  & 0.0046  & 0.9074  & 0.8074  & 0.0358  & 0.3846    \\ 
 & {\color{green}+133\%} & {\color{green}+25\%} & {\color{green}+48\%}  & {\color{green}+148\%} & {\color{green}+2\%} & {\color{green}+1\%} \\

\bottomrule[0.1pt]
\bottomrule[0.1pt]
\end{tabular}
}

        \vspace{-5pt}
\caption{Experimental results on different scenes of the Crello dataset.}
    \label{tab:diffcult}
        \vspace{-0.9cm}
\end{center}
\end{table}

\section{Conclusion}
In this paper, we devise SEGA, a novel stepwise evolution paradigm based content-aware layout generation approach. 
Considering existing methods suffer from significant performance degradation when tackling input with numerous layout elements, we resolve this problem from two perspectives. Firstly, we adopt a coarse-to-fine strategy to break down the tough task. Secondly, we integrate the prior design knowledge into our model to guide the iteratively evaluation and refinement. Moreover, we present a new large-scale poster dataset called GenPoster-100K to enhance the generalization ability of model on unseen domains, which shows potential to facilitate advancements in related fields. Extensive experiments demonstrate the superiority and effectiveness of our proposed approach.

{
    \small
    \bibliographystyle{ieeenat_fullname}
    \bibliography{main}
}

\end{document}